\documentclass[10pt,twocolumn,letterpaper]{article}

\usepackage[pagenumbers]{iccv} 

\definecolor{iccvblue}{rgb}{0.21,0.49,0.74}
\usepackage[pagebackref,breaklinks,colorlinks,allcolors=iccvblue]{hyperref}
\def\paperID{*****} 
\def\confName{ICCV}
\def\confYear{2025}

\title{ObjectVLA: End-to-End Open-World Object Manipulation \\Without Demonstration}

\begin{document}
\author{
Minjie Zhu$^{12*}$\quad
Yichen Zhu$^{1*\dagger}$\quad
Jinming Li$^{3}$ \quad
Zhongyi Zhou$^{2}$\quad
\\
Junjie Wen$^{2}$\quad
Xiaoyu Liu$^{3}$\quad
Chaomin Shen$^{2}$ \quad
Yaxin Peng$^{3}$ \quad
Feifei Feng$^{1}$
\vspace{0.03in}
\\
$^1$Midea Group \quad $^2$East China Normal University \quad $^3$Shanghai University \\
\quad$^*$Equal contribution \quad$\dagger$Corresponding author\vspace{0.1in} \\
\href{https://objectvla.github.io}{\color{blue}\textbf{objectvla.github.io}\xspace}\vspace{-0.3in}
}

\makeatletter
\let\@oldmaketitle\@maketitle%
\renewcommand{\@maketitle}{\@oldmaketitle
    \begin{center}
        \captionsetup{type=figure}
        \centering
    \includegraphics[width=\textwidth]{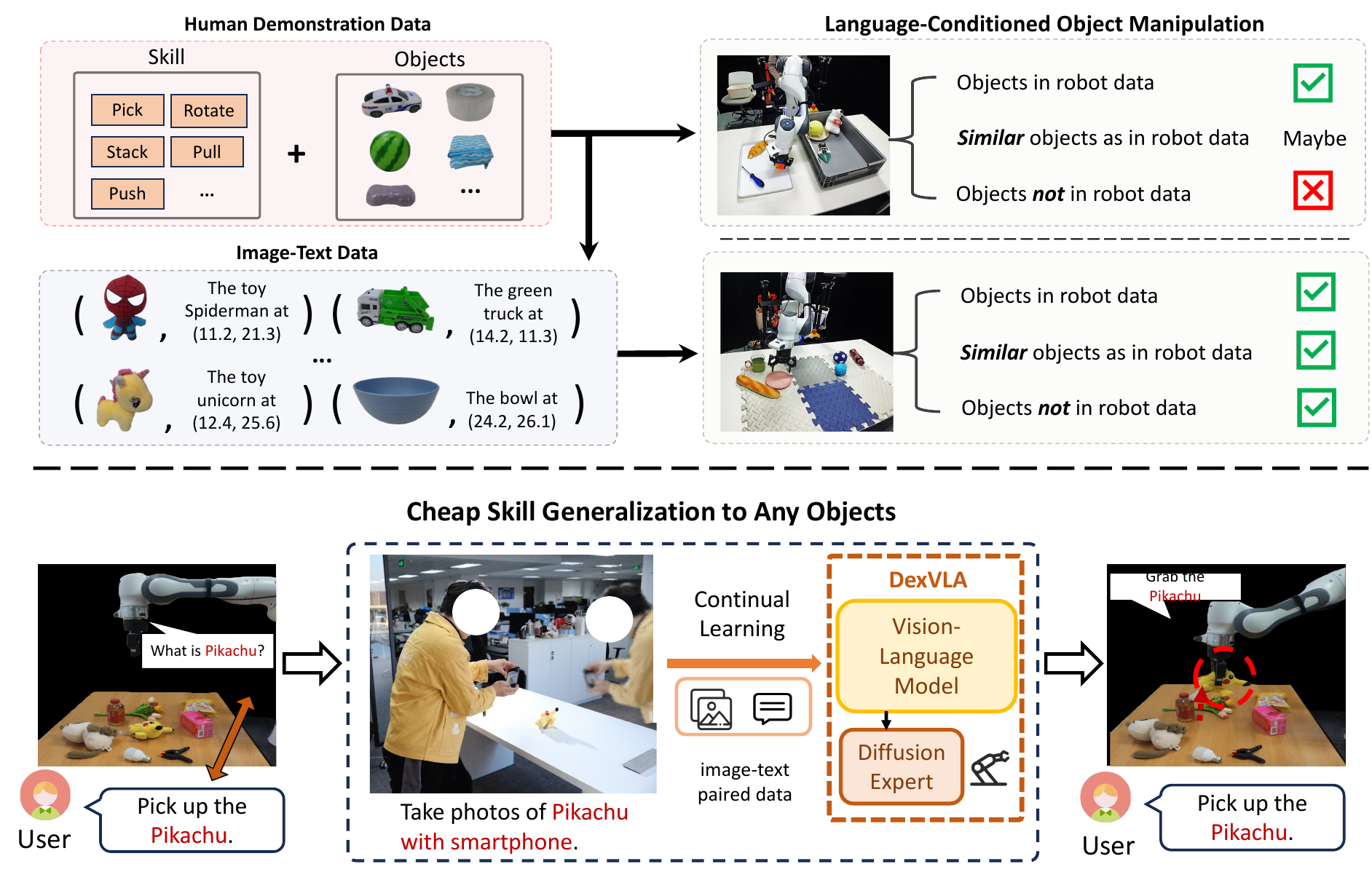}
    \caption{\textbf{A brief illustration of ObjectVLA.} Conventional imitation learning focuses on tasks that involve both skills and objects. While it performs well on seen objects and sometimes generalizes to similar ones (e.g., objects with changed colors), it typically fails with novel objects. By co-training with image-text data, our approach enables VLA models to generalize to any object present in the image-text dataset. \textbf{\textit{Additionally}}, users can capture object images, automatically generate image-text data, and fine-tune a pre-trained VLA model with minimal resources to learn manipulation on novel objects.}\label{fig:robot_main}
    \end{center}
}
\makeatother
\maketitle


\begin{abstract}
Imitation learning has proven to be highly effective in teaching robots dexterous manipulation skills. However, it typically relies on large amounts of human demonstration data, which limits its scalability and applicability in dynamic, real-world environments. One key challenge in this context is object generalization—where a robot trained to perform a task with one object, such as ``hand over the apple,'' struggles to transfer its skills to a semantically similar but visually different object, such as ``hand over the peach.'' This gap in generalization to new objects beyond those in the same category has yet to be adequately addressed in previous work on end-to-end visuomotor policy learning. In this paper, we present a simple yet effective approach for achieving object generalization through Vision-Language-Action (VLA) models, referred to as \textbf{ObjectVLA}. Our model enables robots to generalize learned skills to novel objects without requiring explicit human demonstrations for each new target object. By leveraging vision-language pair data, our method provides a lightweight and scalable way to inject knowledge about the target object, establishing an implicit link between the object and the desired action. We evaluate ObjectVLA on a real robotic platform, demonstrating its ability to generalize across 100 novel objects with a 64\% success rate in selecting objects not seen during training. Furthermore, we propose a more accessible method for enhancing object generalization in VLA models—using a smartphone to capture a few images and fine-tune the pre-trained model. These results highlight the effectiveness of our approach in enabling object-level generalization and reducing the need for extensive human demonstrations, paving the way for more flexible and scalable robotic learning systems.
\end{abstract}

\section{Introduction}

Vision-language-action (VLA) models have emerged as a transformative paradigm for teaching robots dexterous skills, enabling them to replicate human behavior and master complex tasks~\cite{brohan2022rt-1,rt-2, [pi0, pertsch2025fast,ecot}. However, a critical limitation persists: these models rely heavily on human demonstration data, which constrains their scalability and practicality in dynamic real-world environments~\cite{kim24openvla,wen2024tinyvla,wen2024diffusionvla}. For instance, a robot trained to execute “hand over the apple” often fails to generalize to analogous tasks like “hand over the peach,” despite conceptual similarity. This underscores the unresolved challenge of \textbf{object generalization} — adapting learned skills to novel, unseen objects — particularly when such objects lie \textbf{beyond the category of the teleoperated training data}. We name these objects as out-of-distribution (OOD) objects.

The core limitation stems from imitation learning’s tendency to learn fixed mappings from instruction and visual input to action. When encountering objects absent from teleoperation data, the model lacks mechanisms to associate the object’s name, visual features, and learned actions. To address this, we propose a framework that bridges visual-language semantics and robotic actions through localization-aware reasoning.

Our approach begins by curating a dataset of image-text pairs augmented with localization metadata (e.g., bounding boxes). This dataset is co-finetuned with teleoperated robot interaction data, while the robot data itself is enriched with localization-guided reasoning. By embedding localization as a bridging representation, we create a unified pathway between visual-language inputs and robotic actions. This enables zero-shot object generalization: the model can recognize and manipulate novel objects—even those absent from robot training data—without task-specific retraining.

We designed rigorous real-robot experiments to validate the generalization capabilities of our framework, ObjectVLA. In these trials, six objects are positioned at distinct locations (left or right side of a table), with configurations spanning combinations of objects seen in robot interaction data or vision-language data. The robot is tasked with the instruction “move to the {object}”, achieving a 100\% success rate for in-domain objects. To stress-test generalization, we evaluated 100 OOD objects, observing a 64\% success rate. These experiments demonstrate that our method adapts to diverse novel object types when trained with vision-language priors. 

The versatility of our approach is further demonstrated across diverse scenarios, including bin-picking and tasks requiring composite skills like pushing and rotating. Notably, our framework supports rapid adaptation to novel objects: by collecting smartphone-captured images and performing lightweight fine-tuning, the model generalizes to objects absent from the original dataset. These experiments underscore our method’s ability to reduce reliance on large-scale human demonstrations while achieving robust object generalization.

Our primary contribution is a unified pipeline for integrating vision-language datasets with robot interaction data, enabling end-to-end object generalization. Through systematic evaluation, we validate the framework’s performance on complex multi-stage tasks (e.g., bin-picking) and multi-skill manipulation (e.g., rotating, pushing), highlighting its universality. Despite some of the existing works, such as RT-2~\cite{rt-2} and ECoT~\cite{ecot} giving a glimpse of how co-finetuning can achieve simple object generalization, they neither elucidate the underlying mechanism of achieving such generalization nor address the boundary of their methodologies. In contrast, our approach — though simple and straightforward — demonstrates that training VLA models with a hybrid dataset of robot interaction data and image-text data significantly enhances generalization. This level of generalization goes significantly beyond previously demonstrated end-to-end approaches. Crucially, our framework enables practical deployment: even a small set of smartphone images and brief fine-tuning suffices to adapt the model to novel objects, significantly advancing real-world robotic flexibility.

\section{Related Work}

\textbf{Vision-language-action models for robot control.} Recent research has focused on developing generalist robot policies trained on increasingly expansive robot learning datasets~\cite{o2023open-x,khazatsky2024droid,fang2023rh20t,dasari2024ditpolicy,lin2024datascalinglawsimitation}. Vision-language-action models (VLAs) represent a promising approach for training such generalist policies~\cite{kim24openvla,embodiedcot,[pi0,wen2025dexvla,pertsch2025fast,niu2024llarva,diffusion-policy,zhou2025chatvla}. VLAs adapt vision-language models (VLMs)~\cite{zhu2024mipha,zhao2024cobra,zhu2024llavaphi,karamcheti2024prismatic,wang2024qwen2,lu2024deepseek-vl,llava,llava1.5,abdin2024phi3,chen2024internvl}, pre-trained on vast internet-scale image and text data, for robotic control. This approach offers several advantages: leveraging large vision-language model backbones, with billions of parameters, provides the necessary capacity for fitting extensive robot datasets. Furthermore, reusing weights pre-trained on internet-scale data enhances the ability of VLAs to interpret diverse language commands and generalize to novel objects and environments. However, current VLA models struggle to recognize open-world objects when these objects absent from the robot interaction data~\cite{kim24openvla,wen2024tinyvla}. This is mainly due to VLMs essentially “overwrites” its previously acquired knowledge of open-world objects with robot-specific information.

\noindent
\textbf{Generalization in robot learning.} 
In the realm of robot learning, generalization, particularly object generalization, remains a core challenge and active area of research. Many works leverage techniques such as domain randomization~\cite{james2019sim}, meta-learning~\cite{9340848,kaushik2020fast}, retrieval-augmented generation~\cite{zhu2024retrieval}, extra modality~\cite{zhu2025any2policy,ye2025video2policy}, and data augmentation to improve a robot's ability to recognize and interact with novel objects unseen during training. For instance, domain randomization methods~\cite{tobin, james2019sim} randomize visual and physical parameters during simulation training to force the agent to learn features invariant to these irrelevant details, leading to better real-world generalization.  Furthermore, meta-learning approaches~\cite{finn2017model} aim to train models that can rapidly adapt to new objects with limited data, directly addressing the object generalization problem. Finally, data augmentation methods~\cite{laskin2020reinforcement, kostrikov2020image}, enhance the diversity of the training data, exposing the model to a wider range of object appearances and orientations, thereby promoting robustness and generalization to novel objects. There is also a field of work using large language models or vision-language models to do open-vocabulary manipulation~\cite{stone2023open,zhu2024vision,codeaspolicy,ahn2022saycan}, combined with motion planning and robot learning methods. However, these approaches involve separate modules that are trained independently for different components. To the best of our knowledge, this work represents the first exploration of object generalization beyond specific categories within visuomotor policy learning.

\begin{figure*}[t]
    \centering
    \includegraphics[width=\textwidth]{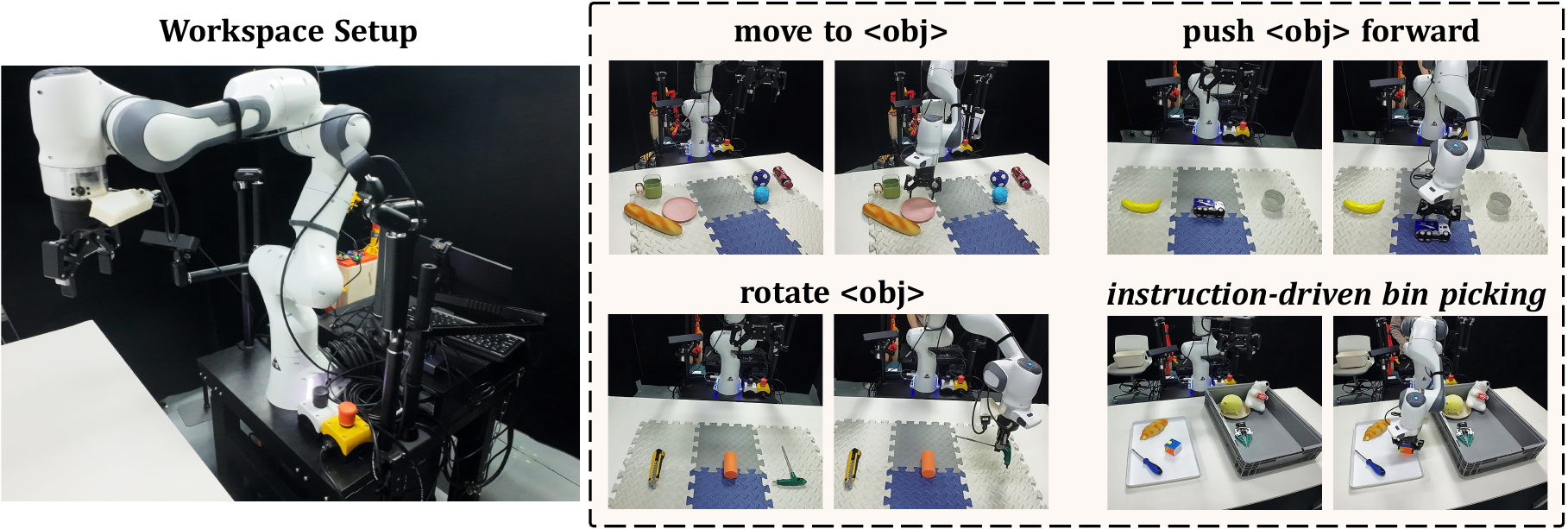}
    \caption{\textbf{Robot setup and examples for real-world manipulation tasks.} We evaluate ObjectVLA with 4 skills on a Franka robot arm equipped with two external Zed cameras and a Realsense 435i wrist camera.}\label{fig:robot_setup}
\end{figure*}

\section{Methodology}
\subsection{Notation and Motivation}
Given a set of expert demonstrations that contain complex robot skill trajectories, we want to learn a visuomotor policy $\pi : \{\mathcal{O}_{r}, \mathcal{I}_{r}\} \mapsto \mathcal{A}$ that maps the visual observations $o_{r} \in \mathcal{O}_{r}$ and the language instruction $i_{r} \in \mathcal{I}_{r}$ to actions $a \in \mathcal{A}$. The action changes accordingly when the language instruction and visual input change. The $r$ denote the data in the human demonstration data. Typically, for each language instruction it contains robot skill such as "push" or "pick up" and the target object, which is denoted as $\{obj_{r}, skill_{r}\} \in i_{r}$. We then formally define the image-text data, where $\varphi : \{\mathcal{O}_{v}, \mathcal{I}_{v}\} \mapsto \mathcal{L}_{v}$, where we input the image $o_{v} \in \mathcal{O}_{v}$ and give a language instruction $i_{v} \in \mathcal{I}_{v}$, the model is output with the corresponding answer $l_{r} \in \mathcal{L}_{v}$. The notation $v$ denotes image-text data.

In this work, we explore the generalization of objects, focusing on those that are not part of the robot interaction data but are present in image-text data.

\subsection{Data Construction}

\textbf{image-text data construction.} To explore the model's ability to generalize to novel objects, we constructed a diverse image-text dataset. For the visual component, we collected 100 distinct objects that are not included in the robot interaction objects. Specifically, using three cameras mounted on the robot (see Figure~\ref{fig:robot_setup}), we captured 20 images per object, covering various poses and orientations to ensure diversity. For the textual component, we employ a fixed template, ``Detecting the bounding box of object.'', as the question, and the corresponding bounding box as the answer. In total, our vision-language dataset comprises 2,000 image-text pairs.

\noindent
\textbf{Reasoning data construction.}
We utilize localization metadata to bridge the gap between image-text data and robot data, as previously mentioned. To establish this implicit link between image-text and action, we incorporate localization metadata into the robot data. This section details how we construct reasoning with localization for robot data.

\begin{figure*}[t]
    \centering
    \includegraphics[width=\textwidth]{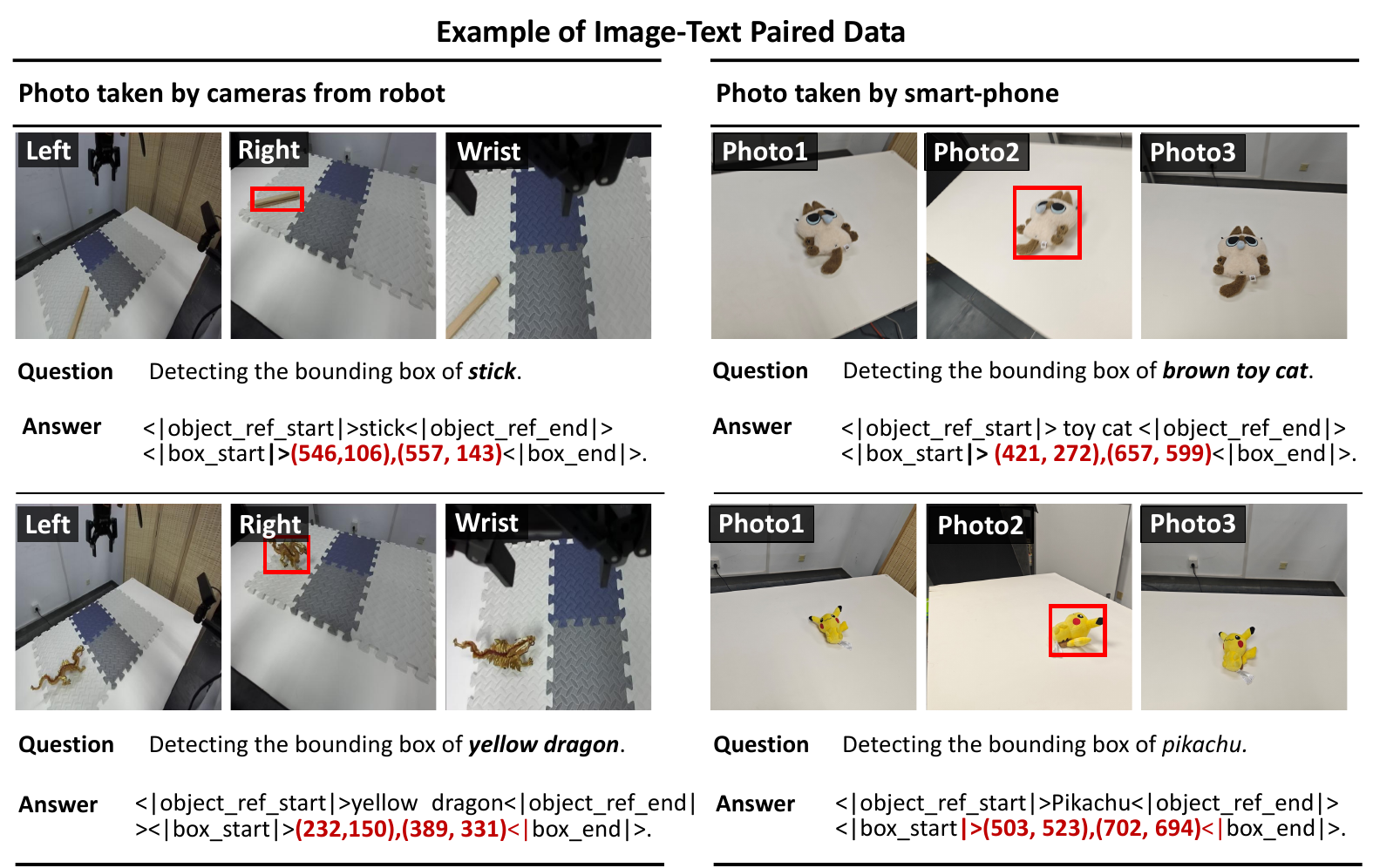}
    \caption{\textbf{Example of constructed image-text data.} \textit{Left:} Photo taken by the robot's camera. \textit{Right:} Object captured with a smartphone.}\label{fig:examples}
\end{figure*}

For each task, we first identify target objects based on the language instructions. We then employ DinoX~\cite{ren2024dino}, a cutting-edge open-vocabulary object detector, to annotate the bounding boxes of these objects. DinoX can generate a bounding box given an object's name. To ensure accuracy, we manually verify and correct any erroneous bounding boxes produced by DinoX. Since our workspace has two external camera views, which can result in different bounding boxes for the same object, we annotate only one (right camera in our experiments). Following Qwen2-VL~\cite{wang2024qwen2}, we use a fixed template, ``$<$$|$object\_ref\_start$|$$>$\{object\}$<$$|$object\_ref\_end$|$$>$$<$$|$box\_start$|$$>$
$(x_{1}, y_{1})$,$(x_{2}, y_{2})$$<$$|$box\_end$|$$>$.'', to represent the localization reasoning. 
This reasoning is generated before each action and injected into the policy model through a learnable module. For a detailed explanation of this injection module's architecture, we refer readers to DiVLA~\cite{wen2024diffusionvla}, the base model used in our experiments. An example of constructed image-text data is at Figure~\ref{fig:examples}.

\subsection{Training Strategy and Implementation Details}
For our experiments, we adopt diffusion-based VLA, a widely used class of Vision-Language-Action (VLA) models exemplified by methods like $\pi_{0}$~\cite{[pi0} and TinyVLA~\cite{wen2024tinyvla}. We select diffusion-based VLA over auto-regressive alternatives due to its significantly faster inference speed, a critical advantage for real-time robotic applications (see FAST~\cite{pertsch2025fast} for a detailed comparison). Specifically, we utilize DiVLA~\cite{wen2024diffusionvla}, a representative VLA architecture, co-train on a hybrid dataset comprising robot interaction data and the vision-language corpus. To balance task-specific adaptation and semantic generalization, we maintained a 10:1 data ratio (robot-to-image-text data) across all tasks. This ratio empirically proved sufficient for robust object generalization, aligning with prior findings on the benefits of co-training for VLA capabilities. Notably, increasing the proportion of robot data beyond this ratio led to a decline in in-domain task success rates. We hypothesize this stems from the limited capacity of the 2B-parameter DiVLA model compared to larger architectures like ECoT (7B)~\cite{ecot} and RT-2 (55B)~\cite{rt-2}, which can better absorb domain-specific data without overfitting.

\section{Experiments}
In this section, we examine the effectiveness of ObjectVLA for object generalization in embodied control. In section~\ref{sec:validate}, we verify the effectiveness of our method in object generalization. In section~\ref{sec:skills} and~\ref{sec:bin-picking}, we illustrate how our model transfers skills to objects not present in robot interaction data but included in the vision-language corpus. In section~\ref{sec:smart-phone}, we show that even a small set of smartphone images and brief fine-tuning can effectively adapt the pretrained model to novel objects.

\noindent
\textbf{Real robot setup.} All experiments are conducted on a Franka robot~\cite{haddadin2024franka} equipped with a 7-degree-of-freedom arm and a gripper. We use two external ZED cameras and a wrist Realsense 435i camera to obtain real-world visual information. Our real-world robot setup is illustrated in Figure~\ref{fig:robot_setup}.

\subsection{Validating Object Generalization}
\label{sec:validate}
In this section, we conduct rigorous experiments to verify the object generalization capability of our method. We begin by describing the experimental setup and evaluation criteria. Next, we evaluate ObjectVLA on both in-distribution and out-of-distribution objects. Finally, we explore several interesting observations related to object generalization.
\subsubsection{Experimental Setup}
To verify the object generalization capability, we begin with a simple yet effective task, ``Move to the object.''. In this task, we position objects on both sides of the robot, ensuring that each side has at least three objects on the table. The model is required to move toward the target object based on the given instruction. These objects are randomly chosen from a diverse set. For in-distribution (ID) evaluation, objects are only selected from the robot's training data. And, for out-of-distribution evaluation, objects are randomly selected from either the robot's training data or the vision-language data. A complete list of objects from both datasets is provided in the Appendix.

\noindent
\textbf{Evaluation criterion.} We evaluate each object over 4 trials, with the target area's side switching every two trials. We consider the model to have successfully recognized a novel object if and only if it moved toward the target object in all four trials. This criterion ensures that the model cannot achieve success simply by chance.

\noindent
\textbf{Experimental Results.} Figure~\ref{fig:move} presents the real-world experimental results for the "Move" task. Our ObjectVLA achieves a 100\% success rate in ID evaluation. In the stress-test evaluation, our model successfully recognizes 64\% of objects that are not present in the robot interaction data, confirming the effectiveness of co-training robot data with localization metadata.

\noindent
\textbf{Ablation study.} To further understand our method's effectiveness, we conducted an ablation study. We found that object generalization relies heavily on two key factors: first, explicitly linking vision and language to action through bounding boxes. This provides a direct connection between the visual object, its linguistic description, and the required manipulation. Second, a reasoning process for the robot data should be designed that mirrors the structure of vision-language pair data. This allows the model to leverage the rich information encoded in pre-trained vision-language models.

To analyze the impact of these factors, we removed the reasoning module for robot data and eliminated bounding boxes for vision-language data. The VLA model is then co-finetuned with vision-language data and evaluated using the same criteria and test settings as our full method.

As illustrated in Figure~\ref{fig:move}, the model without bounding boxes achieves only a 19\% success rate in OOD evaluation, representing a significant performance decline compared to our method, despite achieving a 100\% success rate in the ID test. This suggests that without explicit grounding and a structured reasoning process, the model struggles to differentiate objects in vision-language data, leading to confusion about object-instruction correspondence and appropriate action selection.

\subsubsection{More Observations}

\begin{figure}[t]
    \centering
    \includegraphics[width=0.5\textwidth]{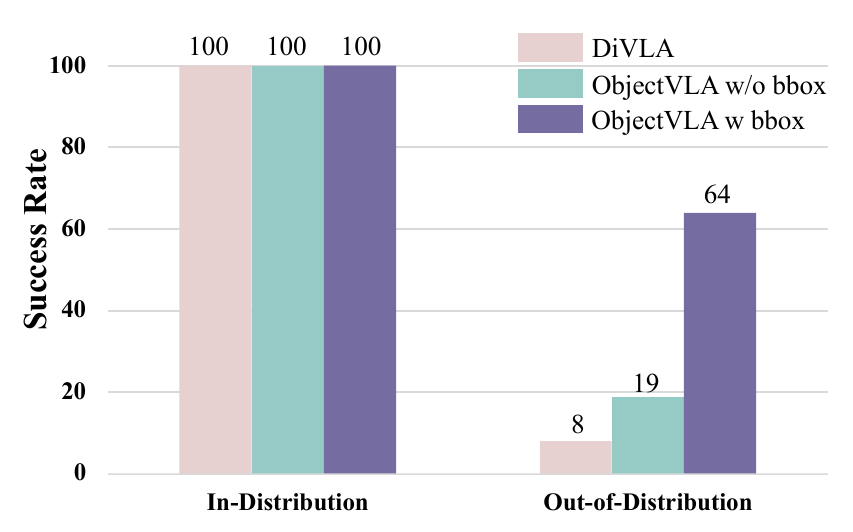}
    \caption{\textbf{Validation experiments on object generalization.} 
Our method achieved the best performance in both the in-distribution test setup and under visual changes. Each object is evaluated across 4 trials. We report the number of objects that were correctly identified in all four trials.}
    \label{fig:move}
\end{figure}

\textbf{Can VLA recognize unseen objects if only trained with teleoperated data?} To further assess the importance of vision-language data, we evaluated a VLA model trained exclusively on robot data, without any vision-language co-finetuning. As shown in Figure~\ref{fig:move}, this model (DiVLA) achieved 8\% accuracy, which is almost equivalent to random guessing. This stark outcome highlights the critical role of vision-language data in multimodal understanding.

While the VLA model's backbone is pre-trained on internet-scale vision-language data, focusing solely on robot data during training leads to catastrophic forgetting. The model essentially ``overwrites'' its previously acquired knowledge of visual concepts with robot-specific information, hindering its ability to comprehend multimodal scenes. Consequently, even objects encountered during pre-training, such as Pikachu, remain unrecognizable to the VLA model without vision-language co-finetuning.

\begin{figure*}[h]
    \centering
    \includegraphics[width=\textwidth]{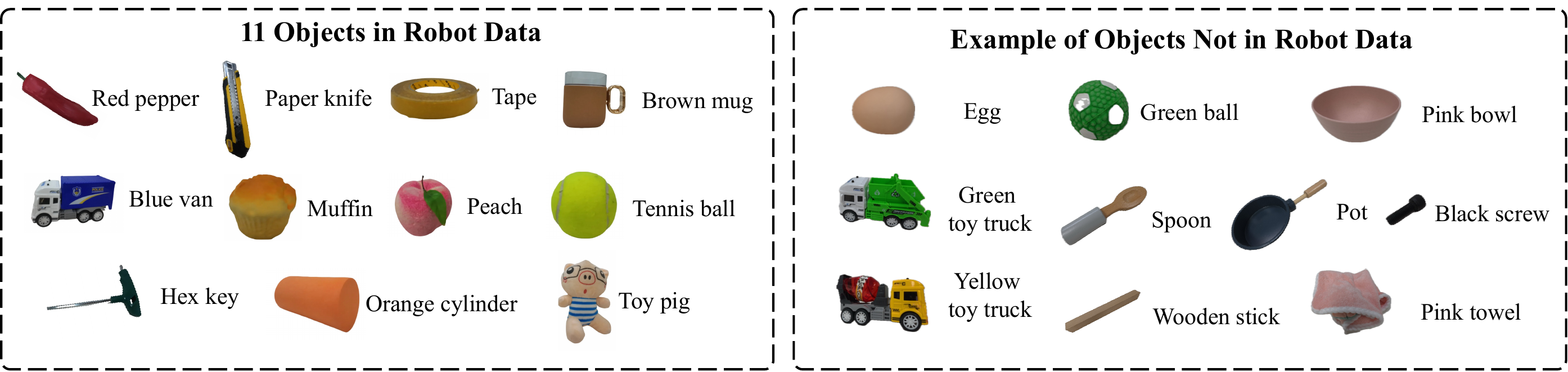}
    \caption{\textbf{Example Objects Used in Experiments.} \textit{Left:} Objects present in the robot training data. \textit{Right:} Examples of novel objects, not present in the robot data, but included in the image-text co-training dataset (see Appendix for a comprehensive list).}\label{fig:seen-objs}
\end{figure*}

\subsection{Combining with More Skills}
\label{sec:skills}
While the previous section employed a simple ``move to'' demonstration to validate the fundamental approach of our method, this section expands the evaluation to encompass more complex skills, specifically "push" and "rotate." This broader assessment aims to demonstrate the generalizability of our method and its applicability beyond the "move to" task.
\begin{table}[t]
\centering
\vspace{0.1cm}
\caption{\textbf{Experimental Results for rotate and push skills.} Our proposed ObjectVLA achieves high performance on both 5 in-distribution objects and 20 out-of-distribution objects. Each object is evaluated with three trials. We report the number of success trials.}
\label{tab:rotate}
\resizebox{0.45\textwidth}{!}{\begin{tabular}{c|c|c}
\toprule
\textbf{Task}  & In-Distribution & Out-of-Distribution \\
\midrule
Rotate & 13/15  & 39/60    \\
Push   & 12/15  & 52/60    \\ 
\bottomrule
\end{tabular}}
\vspace{-0.5cm}
\end{table}

\noindent
\textbf{Experimental setup.} In this experimental setup, we placed three objects in front of the robot: one on the center, one on the right, and one on the left. The robot is instructed to either "rotate the object counterclockwise" or "push the object forward," as illustrated in Figure~\ref{fig:robot_setup}. Following previous setup, we evaluate the model's performance for both in-distribution (ID) and out-of-distribution (OOD) objects. Recognizing that some objects are inherently unsuitable for rotation or pushing actions (e.g., dishes), we conducted experiments on a curated set of 5 ID objects and 20 OOD objects. For each object in a skill, 40 demonstrations were collected, resulting in a total of 400 demonstrations.

\noindent
\textbf{Implementation details.} We train one model for each skill to ensure that the model focuses more on understanding the objects rather than multi-task learning. 
We use the same image-text data of ``move'' task. Following established protocols from our prior work, this image-text dataset trained concurrently with the demonstration data for comprehensive evaluation. Each object was tested with 3 trials. In total, 150 trials were conducted. The training setting is provided in Appendix.

\noindent
\textbf{Results.} As shown in Table~\ref{tab:rotate}, our method achieved high success rates on the robot interaction objects for both rotate and push skills. Analysis of the failed ``rotate'' trials revealed that the primary cause is the model's inability to grasp the target object securely. When evaluating performance on out-of-distribution (OOD) objects, we observed a decrease in task completion rates compared to in-distribution objects, as expected. However, the model still successfully completed nearly two-thirds of the trials. Notably, in most failure cases, the model did not incorrectly identify the target object but rather failed to execute the skill completely. This was particularly evident in the ``rotate'' trials, where successful execution hinges on a secure grasp, a challenging requirement for unseen objects. Nevertheless, these experiments strongly support the claim that ObjectVLA can transfer learned skills, beyond basic pick and place, to novel objects within the framework we have developed. The results underscore the potential of ObjectVLA for generalized robotic manipulation, capable of adapting to new objects and tasks beyond its initial training.

\subsection{Instruction-Driven Bin Picking}
\label{sec:bin-picking}
\begin{table}[t]
\centering
\vspace{0.1cm}
\caption{\textbf{Experimental results for bin picking.} Our proposed ObjectVLA achieves high performance on both 11 in-distribution objects and 50 out-of-distribution objects, with each object evaluated across 3 trials. We report the number of successful trials over total trials.}
\label{tab:bin-picking}
\resizebox{0.45\textwidth}{!}{\begin{tabular}{c|c|c}
\toprule
\textbf{Method}  & In-Distribution & Out-of-Distribution \\
\midrule
OpenVLA    & 14/33  &  17/150 \\
ObjectVLA &  \textbf{21/33}  &  \textbf{87/150}\\
\bottomrule
\end{tabular}}
\vspace{-0.5cm}
\end{table}
To further evaluate ObjectVLA, we conducted experiments in a more practical scenario: end-to-end instruction-driven bin-picking. Unlike prior works (e.g., GR-2~\cite{cheang2024gr2} and DiVLA~\cite{wen2024diffusionvla}) that execute bin-picking tasks without specific semantic instructions—typically limited to generic actions like transferring all objects from one container to another—we focus on a significantly more challenging setting~\cite{cheang2024gr2,wen2024diffusionvla}. In our experiments, the robot is required to identify and retrieve a specific target object based on natural language instructions (e.g., "Pick the hexagonal bolt from the bin"). This scenario elevates the complexity of conventional bin-picking tasks by integrating cross-modal understanding (vision-to-language alignment) and fine-grained object discrimination that have multiple objects in the scene. Notably, the objects are randomly placed on the panel, which is a large area. Not only does the model need to figure out the object's position, but also needs to be aware of its pose. 

\noindent
\textbf{Implementation details.} We collected new data within this environment.  For robot interaction data, we collected 600 pick-and-place trajectories using the same "seen" objects as in previous experiments.  For image-text data, we used half the number of objects from previous experiments, capturing 20 images of each. We compared our method against OpenVLA, a state-of-the-art VLA model, reporting success rates for both in-distribution and out-of-distribution objects. Evaluation consisted of three trials per object, totaling 183 trials per method. In each trial, at least two objects were randomly placed on the plate, and the model was instructed to pick and place a specific object according to the given instruction.

\noindent
\textbf{Results.} Table \ref{tab:bin-picking} presents our experimental results.  Bin picking, requiring object retrieval from random positions and poses, poses a significant challenge even for in-distribution objects.  OpenVLA achieves a success rate of only 42.4\% for in-distribution objects, significantly less than half. Surprisingly, it still completed roughly 10\% of trials with out-of-distribution objects. This is likely due to some test objects sharing attributes with training objects (e.g., bread resembling a muffin, a green mug differing only in color from a brown training mug).  In contrast, our method successfully completed 87 of 150 trials, including many completely novel objects, a 46.7\% improvement over OpenVLA. This further emphasizes the necessity of co-training with both robot interaction and image-text data for effective object generalization.

\begin{figure}[t]
    \centering
    \includegraphics[width=0.35\textwidth]{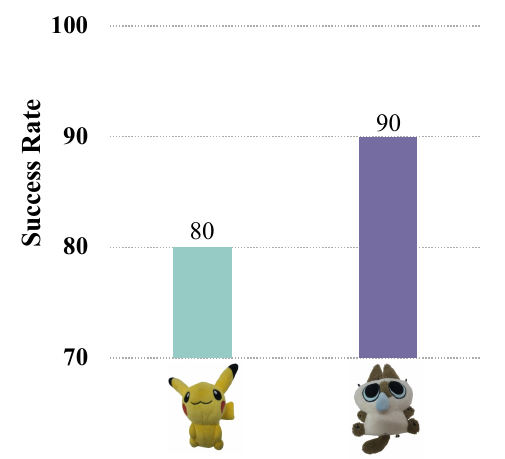}
    \caption{\textbf{Experimental results for smartphone captured objects and trained by continual learning.} We test two new objects. We took pictures of these two objects via smartphone and continually trained them on a pre-trained model. Each object was evaluated across 10 trials. We report the success rate for each object.}\label{fig:smart-phone}
\end{figure}

\subsection{Cheap Object Generalization via Smart-Phone Pictures and Continual Learning}
\label{sec:smart-phone}

The previous section demonstrated that our proposed co-training strategy enables the imitation learning method to generalize to any object by constructing corresponding image-text data for each object. This approach significantly enhances the model's ability to handle a broader range of objects, without requiring extensive retraining. However, there are two key limitations that need to be addressed.

First, when a new object is introduced, the model must be trained from scratch to incorporate the new object into its understanding. This process can be highly cost-inefficient, as it involves retraining the model every time a novel object is added. In real-world applications, where objects are frequently introduced or changed, this limitation could significantly slow down deployment and increase operational costs. Second, our current image data is collected using cameras mounted on the robot, which ensures that there is no visual gap between the images captured by the robot's cameras and the images input into the model. This setup works well in controlled environments but presents challenges in real-world scenarios. For instance, in order to capture the same images as the robot sees, you would need to replicate the exact camera positions and angles of the robot’s setup. This is not only cumbersome but also expensive, as it requires building an identical system with matching camera views. Moreover, in environments where the robot is mobile or the scene is dynamic, maintaining consistent camera alignment becomes even more difficult.

Therefore, in this section, we test a simpler and more cost-effective approach: using a smart-phone camera to collect images from various perspectives. The model is then continuously trained on the pre-trained weights using this more accessible data collection method. As shown in Figure~\ref{fig:smart-phone}, we test with two objects: Pikachu and a brown toy cat. For each object, we capture 21 images and follow the same data construction pipeline discussed earlier. We train these objects in a bin-picking environment. Our results demonstrate that the model is able to recognize and successfully grasp the objects with a high success rate, 80\% for Pikachu and 90\% success rate for the toy cat. More importantly, we only need to continue training the model for 1 epoch. Because the collected data size is small, the training process can be extremely fast and can be finished up to ten minutes. This validates the effectiveness of our approach, showing that simple smartphone image collection combined with continuous learning enables open-world object manipulation in an end-to-end model. This experiment demonstrates that our method is flexible and cost-effective, making it a plug-and-play solution for existing VLA models, enabling them to generalize to virtually any object.

\section{Conclusion}
In this work, we present ObjectVLA, a Vision-Language-Action framework that addresses object generalization in robotic manipulation. By integrating vision-language datasets with robot interaction data, our method establishes a unified pipeline that bridges semantic understanding and physical action execution. This enables zero-shot generalization to over 100 novel objects with a 64\% success rate, even when objects differ in category, appearance, or fine-grained attributes (e.g., color, shape). Our framework demonstrates that lightweight co-training with image-text priors and localization-aware reasoning can unlock robust cross-modal alignment. Key to our success is the ability to adapt rapidly to real-world scenarios: using just a few smartphone-captured images and quick continual finetuning, robots generalize to unseen objects without costly human demonstrations. We validate our approach across diverse tasks—including bin-picking, rotating and pushing—showcasing its versatility and practicality. Our results highlight a path toward scalable robotic learning systems that reduce dependence on large-scale teleoperation data while maintaining high performance.

\section{Limitation}
There are still a number of limitations in this work. Specifically, the image-text data was collected by the authors using either a robot-mounted camera or a smartphone. While we have not yet explored the feasibility of leveraging internet-sourced image-text data, it presents an intriguing avenue for future research.  Specifically,  investigating the necessary degree of visual similarity between internet images and target objects for effective skill transfer would be valuable. Our primary focus here is to introduce a novel pipeline that enables deep learning models to transfer skills to new objects without explicit demonstrations.  Determining the limits of this transferability, particularly concerning the permissible visual gap between training and target objects, remains an open question for future investigation. Currently, our method struggles to generalize to novel backgrounds and lighting conditions. We believe the visual gap between our collected image-text data and the robot's operational environment contributes to this challenge.  Bridging this gap to improve generalization is a key focus for future development.
{
    \small
    \bibliographystyle{ieeenat_fullname}
    \bibliography{main}
}

\begin{figure*}[h]
    \centering
    \includegraphics[width=0.95\textwidth]{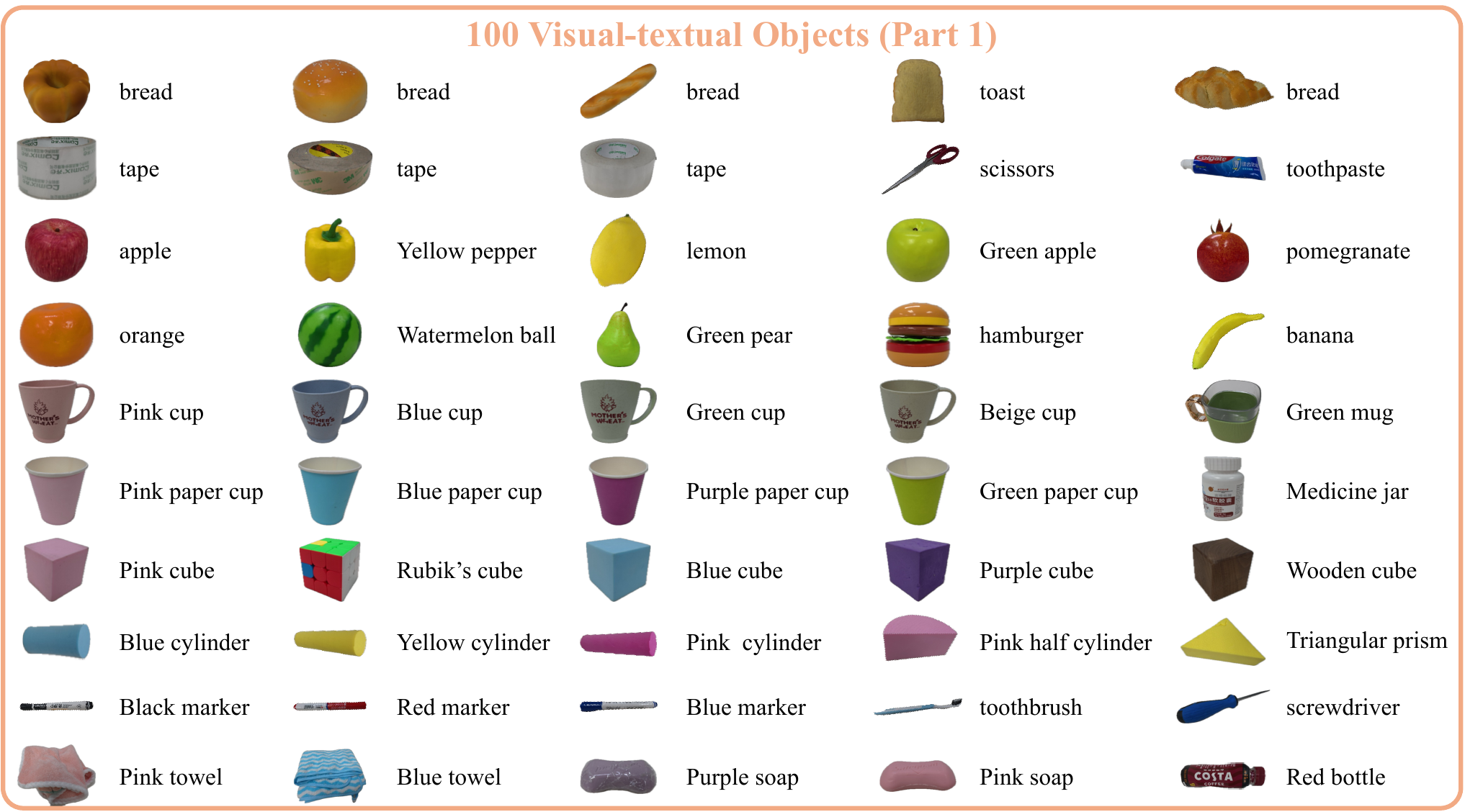}
    \caption{The out-of-distribution objects used in our experiments (part 1).}\label{fig:ood_objs_part1}
\end{figure*}

\section{Appendix}
\label{sec:appendix}

\subsection{Evaluation Metrics}
For real robot task, we record the percentage of trials where the robot successfully completes the assigned task. This is a fundamental metric for any robot experiment. Multiple trials are conducted for the evaluation.

\subsection{Implementation Details}
All experiments were conducted on eight NVIDIA A800 GPUs using the Adam optimizer with a constant learning rate of 2e-5 and a global batch size of 128.  Training proceeded for 50,000 steps, with the final checkpoint selected based on validation performance. Unless otherwise stated, the ratio of robot data to visual-text data was 10:1.  Empirically, we observed that increasing the proportion of robot data significantly degraded manipulation performance.  Our base model is DiVLA\cite{wen2024diffusionvla} with a Qwen2-VL-2B backbone~\cite{wang2024qwen2}. As our focus is developing a co-training method for novel object generalization, we retained the original model architecture.

\subsection{Example of Objects Used in Experiments}
\begin{figure*}[t]
    \centering
    \includegraphics[width=0.95\textwidth]{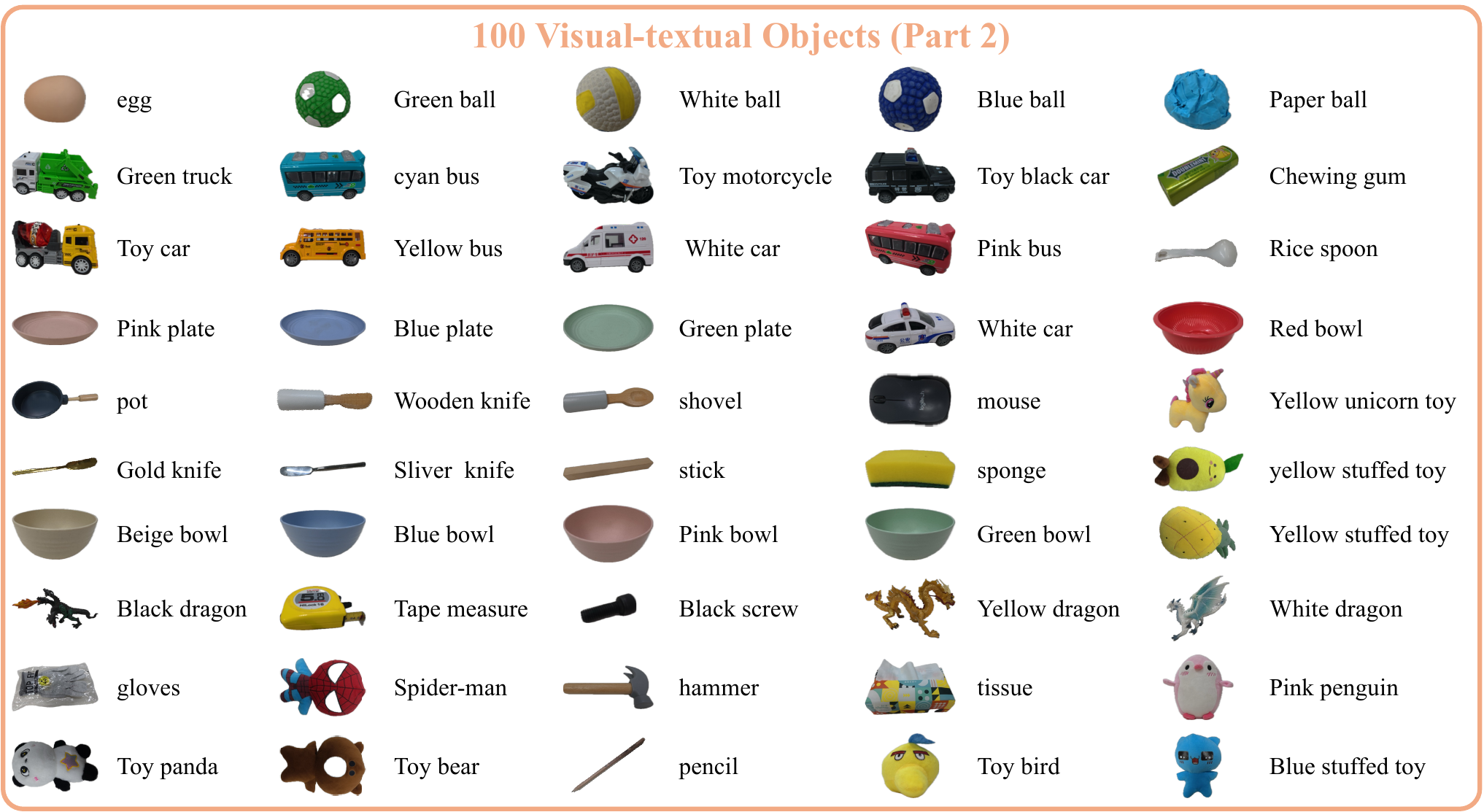}
    \caption{The out-of-distribution objects used in our experiments (part 2).}\label{fig:ood_objs_part2}
\end{figure*}

We provide a comprehensive list of \text{out-of-distrbution} objects and names that we used for training and evaluation, which are shown in Figure~\ref{fig:ood_objs_part1} and Figure~\ref{fig:ood_objs_part2}.

\end{document}